\documentclass[preprint,review,1p,times,sort&compress]{elsarticle}

\usepackage{amssymb}
\usepackage{amsmath}
\usepackage{pifont}

\usepackage{algorithm}

\usepackage{algorithmicx}
\usepackage{algcompatible}

\usepackage{xcolor} 
\usepackage{booktabs}
\usepackage{multirow}
\usepackage{hyperref}

\journal{Pattern Recognition}

\begin{document}

\begin{frontmatter}



\title{CASP: Few-Shot Class-Incremental Learning with CLS Token Attention Steering Prompts} 



\author{Shuai Huang\textsuperscript{a},  
        Xuhan Lin\textsuperscript{a}, 
        Yuwu Lu\textsuperscript{a,*}
        }
\cortext[cor1]{
    School of Artificial Intelligence, South China Normal University, Foshan, 528225, China. \\
    \hspace*{1.5em} \textit{E-mail} address: \href{mailto:luyuwu2008@163.com}{luyuwu2008@163.com} (Y. Lu).
    }





\affiliation[scnu]{organization={School of Artificial Intelligence},
addressline={South China Normal University}, 
city={Foshan},
country={China}}




\begin{abstract}
Few-shot class-incremental learning (FSCIL) presents a core challenge in continual learning, requiring models to rapidly adapt to new classes with very limited samples while mitigating catastrophic forgetting. Recent prompt-based methods, which integrate pretrained backbones with task-specific prompts, have made notable progress. However, under extreme few-shot incremental settings, the model's ability to transfer and generalize becomes critical, and it is thus essential to leverage pretrained knowledge to learn feature representations that can be shared across future categories during the base session. Inspired by the mechanism of the CLS token, which is similar to human attention and progressively filters out task-irrelevant information, we propose the CLS Token Attention Steering Prompts (CASP). This approach introduces class-shared trainable bias parameters into the query, key, and value projections of the CLS token to explicitly modulate the self-attention weights. To further enhance generalization, we also design an attention perturbation strategy and perform Manifold Token Mixup in the shallow feature space, synthesizing potential new class features to improve generalization and reserve the representation capacity for upcoming tasks. Experiments on the CUB200, CIFAR100, and ImageNet-R datasets demonstrate that CASP outperforms state-of-the-art methods in both standard and fine-grained FSCIL settings without requiring fine-tuning during incremental phases and while significantly reducing the parameter overhead. Our code is available at \url{https://github.com/huangshuai0605/CASP}. 
\end{abstract}




\end{frontmatter}



\section{Introduction}
\label{Introduction}
Human brains exhibit unique proficiency in recognizing novel concepts: the ability to effectively learn from very limited examples while retaining previously acquired knowledge. Contemporary research focuses primarily on emulating the human learning systems through two prominent paradigms: few-shot learning (FSL) \cite{PRFSL} and class-incremental learning (CIL) \cite{L2P,DualPrompt,CODA-Prompt,APT,PRCIL}. However, when confronted with the combined challenges of both paradigms in few-shot class-incremental learning (FSCIL) \cite{FewShotCIL,SemanticAwareKD_FSCIL,FILP3D,PanPFR,FewShotClassIncrementalLearning}, the existing methodologies frequently encounter substantial limitations in generalizing to novel concepts or mitigating catastrophic forgetting of foundational knowledge, primarily due to the extreme scarcity of training samples for new classes.

Both CIL and FSCIL confront the fundamental challenge of catastrophic forgetting \cite{French1999}. Furthermore, FSCIL is particularly susceptible to overfitting \cite{FewShotCIL}: models tend to memorize superficial patterns from limited new samples rather than learning transferable and discriminative features, ultimately resulting in performance degradation on both newly acquired and previously learned classes. Recent prompt-based approaches \cite{L2P, DualPrompt,CODA-Prompt}, which maintain a fixed pretrained backbone while acquiring task-specific prompts, have demonstrated notable success in CIL. Nevertheless, their direct application to FSCIL has proven suboptimal. The key-query matching mechanism often underperforms on scarce data during the incremental phases, failing to learn effective prompts for new tasks. Moreover, the computational overhead of prompt pools is too substantial to be compatible with efficient adaptation. While existing studies have established the superiority of additive prompts \cite{APT} over conventional input concatenation methods, this architectural design remains inadequate for FSCIL demands, as it lacks mechanisms to enhance feature generalization and structure the feature space for future tasks.

This challenge has led to a fundamental insight: human cognition demonstrates a selective attention capability \cite{DESIMONE1995} to progressively disregard irrelevant details and concentrate on the most salient features when processing extensive information. This capacity enables rapid identification and acquisition of novel concepts, even under extreme sample constraints. To mimic the sophisticated human learning mechanism, we introduce the CLS Token Attention Steering Prompts (CASP) for FSCIL, a novel framework meticulously designed to meet the stringent FSCIL requirements. CASP integrates three key innovations that transform additive prompting into an efficient and effective paradigm for FSCIL. First, symmetric prompt injection extends the additive mechanism to concurrently modulate the query, key, and value projections of the CLS token, allowing for a more comprehensive recalibration of the self-attention mechanism and enhancing the model's ability to extract and amplify the most discriminative features from few-shot instances. Second, attentional stochastic perturbation introduces a targeted dropout \cite{SRIVASTAVA2014} strategy that randomly perturbs the attention weights of the prompted CLS token, serving as a powerful regularizer that simulates noise and encourages the model to learn more generalized and stable decision boundaries, thereby explicitly counteracting overfitting and enhancing robustness. Finally, Manifold Token Mixup (MTM) is incorporated to foster smoother decision boundaries and a more structured latent feature space, which not only improves generalization but also implicitly reserves capacity for future classes, reducing feature entanglement and preparing the model for sustained incremental learning.

Through comprehensive experiments on standard FSCIL benchmarks (CUB200 \cite{CUB2011}, CIFAR100 \cite{CIFAR100}, and ImageNet-R \cite{Hendrycks2021}), we demonstrate that CASP achieves state-of-the-art performance without requiring fine-tuning during the incremental phase. Our analysis confirms that CASP effectively mitigates forgetting, counteracts overfitting, and learns a well-structured feature space while maintaining high parameter efficiency. In summary, our contributions are as follows:

\begin{itemize}
    \item We propose the CASP framework, a novel framework that implements symmetric prompt injection across query, key, and value projections to enhance feature adaptation from limited samples.

    \item We incorporate two novel regularizers: attentional dropout for improved robustness and Manifold Token Mixup for structured feature space consolidation.

    \item We empirically demonstrate the superior efficacy of CASP through extensive experiments, establishing new state-of-the-art results across multiple benchmarks while requiring fewer parameters and no fine-tuning during incremental phases.
\end{itemize}


The rest of the paper is as follows: 
Section \ref{Related works} introduces related research on few-shot class-incremental learning, prompt-based approaches, and Mixup-based augmentation. 
Section \ref{Background} presents the problem formulation and essential groundwork. 
Section \ref{Method} presents our proposed approach in detail. 
Section \ref{Experiments} outlines the experimental setup and presents the results validating the effectiveness of our proposed method. 
Section \ref{Conclusion} summarizes this work.

\begin{figure}[t]
\centering
\includegraphics[width=1.0\textwidth]{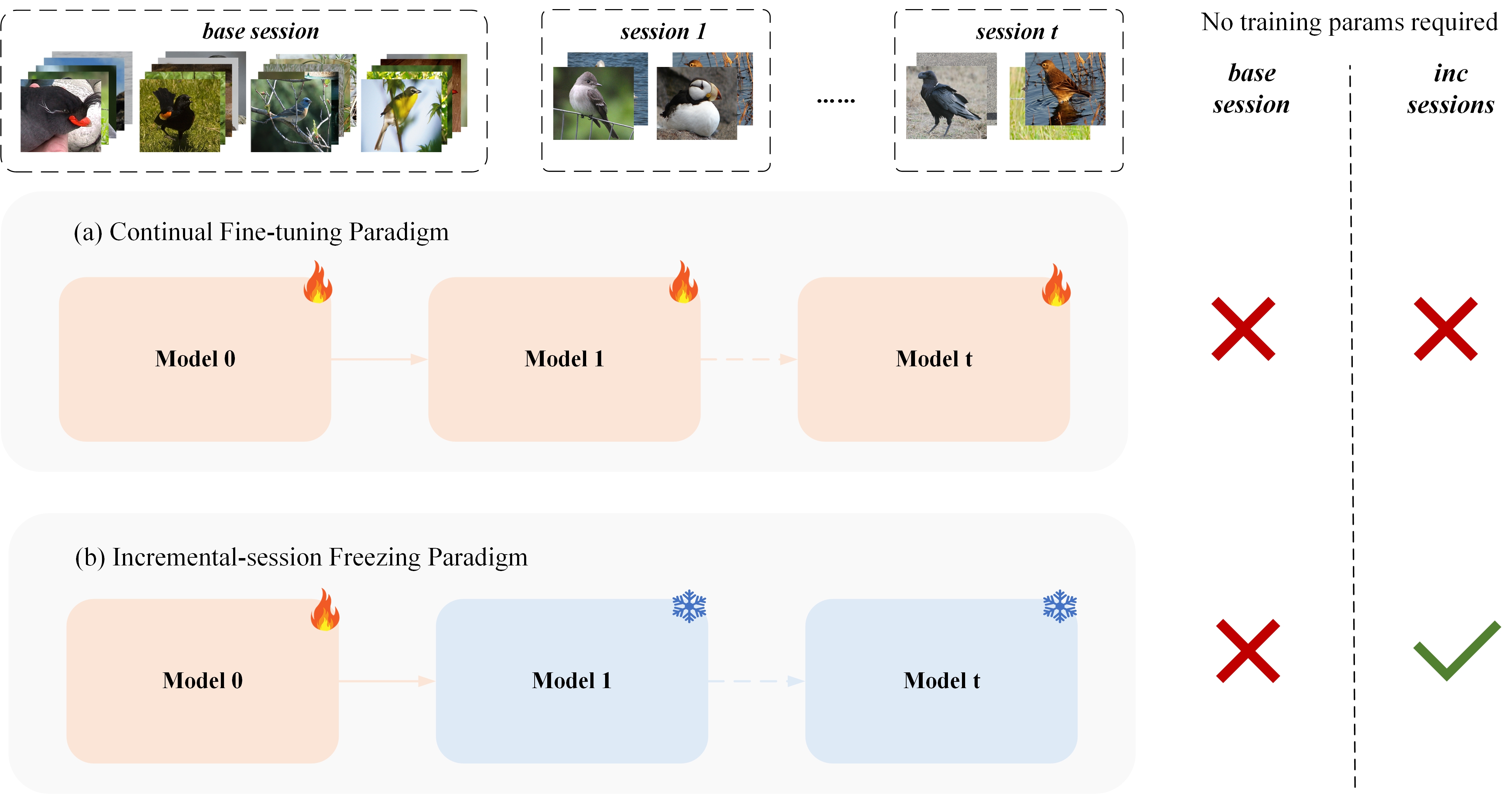}
\caption{Comparison of two continual learning paradigms. In the continual fine-tuning paradigm, the model parameters are updated in each incremental session. In the incremental-session freezing paradigm, only the base session involves model training, while the parameters are frozen in subsequent incremental sessions, requiring no updates.}\label{first}
\end{figure}

\section{Related works}
\label{Related works}

\subsection{Few-shot class-incremental learning}
\label{Few-shot class-incremental learning}
 FSCIL presents a more demanding scenario than conventional class-incremental settings, as it necessitates the continuous integration of novel classes from only a handful of labelled samples per category. The FSCIL paradigm was formally established by TOPIC \cite{FewShotCIL}, and it introduced a standardized benchmark in which a model is initially trained on a comprehensive base class set and subsequently learns a sequence of tasks, each comprising new classes with severely limited data. The existing methodologies in FSCIL can be broadly categorized into two principal research threads. Backwards-compatible approaches \cite{SemanticAwareKD_FSCIL, FewShotClassIncrementalLearning,FlatMinima,EvolvedClassifiers,ClassAwareDistillation,GKEALFramework} primarily aim to mitigate catastrophic forgetting and overfitting caused by parameter adjustments during incremental updates, while in contrast, forwards-compatible methods \cite{OrCo,SAVC,TEEN,NeuralCollapseFSCIL,FACT} emphasize the acquisition of transferable and discriminative representations \cite{BEI2025} from base classes to facilitate generalization towards future tasks. The former methods often suffer from significant forgetting due to extensive backbone fine-tuning, whereas the latter methods, by keeping the backbone frozen, exhibit restricted plasticity, thereby exacerbating the stability-plasticity dilemma. The distinction between these two approaches is illustrated in Fig. \ref{first}.

Recent prompt-based techniques have demonstrated considerable promise in terms of facilitating efficient knowledge transfer and retention in FSCIL. For instance, PKT \cite{PKT} selectively fine-tunes certain layers of the Vision Transformer (ViT) backbone while incorporating prompts to encapsulate both domain-specific and semantic knowledge. Similarly, ASP \cite{ASP} employs a frozen ViT backbone enhanced with both attention-aware task-invariant prompts and learnable task-specific prompts to concurrently alleviate catastrophic forgetting and overfitting. SEC \cite{SEC} further introduces a prompt pool mechanism, explicitly decoupling prompts into discriminative and non-discriminative components. Despite their effectiveness, these methods predominantly operate in the input space, overlooking the pivotal role of the CLS token, which aggregates global contextual information through self-attention by progressively filtering irrelevant details and concentrating on the most salient class-relevant features.

In this work, we advocate for the focused refinement of the CLS token during the base session to steer the model towards learning domain-generalizable representations. Our approach operates in parallel with forwards-compatible strategies and remains architecturally conservative, requiring no structural modifications to the underlying ViT model. By attentively modulating the information aggregation process of the CLS token, we aim to achieve more robust and generalizable few-shot incremental learning.

\subsection{Prompt-based approaches}
\label{Prompt-Based Approaches}
Prompt-based methods were initially proposed in natural language processing (NLP) to better leverage pretrained knowledge in downstream tasks \cite{Chain-of-thought}. The core idea is to keep the backbone parameters frozen and fine-tune only a small number of prompt parameters, which are prepended to the input sequence. In the context of Vision Transformers, prompt tuning \cite{VPT} and prefix tuning \cite{PKT,prefix_tuning} have emerged as widely adopted strategies. Prompt tuning introduces learnable tokens into the input space, whereas prefix tuning appends prompts to the key and value projections within the self-attention mechanism, thereby directly modulating the attention patterns to capture task-specific knowledge.

These techniques have been effectively extended to CIL. For instance, L2P \cite{L2P} employs a key-query matching mechanism to retrieve task-specific prompts from a predefined prompt pool, while DualPrompt \cite{DualPrompt} enhances this approach by incorporating both task-specific and task-invariant prompts, with the latter aiming to capture shared knowledge across tasks. However, these methods still face challenges: task-invariant prompts often retain task-specific information, and the prompt selection mechanism requires ample data during the incremental phases to learn effectively, a condition rarely satisfied in few-shot settings. Consequently, they are prone to overfitting when new task samples are extremely limited, leading to performance degradation in FSCIL.

Recent approaches have sought to dynamically generate prompts to improve adaptability. Coda-prompt \cite{CODA-Prompt} composes input-conditioned prompts by using a set of learnable components weighted by input-dependent attention. APT \cite{APT} uses a set of shared prompts applied during the attention computation of CLS token and updates them gradually via an exponential moving average (EMA), significantly reducing the parameter overhead and mitigating catastrophic forgetting. VFPT \cite{VFPT} integrates fast Fourier transform into prompt embeddings to combine the spatial and frequency domain information, effectively enhancing the representation capability of prompts.

Despite these advancements, existing prompt-based methods remain limited in FSCIL due to their dependency on sufficient data for learning effective prompts. In contrast, our method eliminates the need for task-specific prompt training during incremental sessions, focusing instead on learning domain-generalizable representations in the base session. This approach considerably reduces parameter overhead and improves adaptability to FSCIL scenarios, resulting in stronger generalization and more efficient knowledge retention.

\subsection{Mixup-based augmentation}
\label{Mixup-Based Augmentation}
In the field of data augmentation for deep visual models, interpolation-based methods from the Mixup family have been demonstrated to effectively enhance model generalizability and robustness. Early approaches such as Mixup \cite{mixup2018} and its variant, Manifold Mixup \cite{Manifold}, primarily operate in the pixel or feature space, whereas the rise of the ViT \cite{ViT} has spurred the development of token-level mixing strategies, such as TokenMix \cite{TokenMix2022} and TokenMixup \cite{TokenMixup}, which align well with the patch-based input structure and self-attention mechanisms of ViT. These methods often leverage attention maps or lightweight networks to selectively blend informative tokens, thereby improving both efficiency and semantic consistency. However, key challenges remain in this domain, involving how to design mixing operators that preserve the token structure and how to ensure label semantic consistency under partial token mixing. These open questions indicate promising directions for future research. In this paper, we explore whether manifold mixing of input tokens in ViT can simulate future categories, enhance generalization, and reserve the feature space for unseen classes.

\section{Background}
\label{Background}

\subsection{Problem definition}
\label{Problem definition}
FSCIL aims to enable a model to continually acquire knowledge from a sequence of learning sessions 
$\mathcal{S}_0, \mathcal{S}_1, \ldots, \mathcal{S}_T$, each containing limited labelled examples. During training on session $t$, access is limited to the current dataset $\mathcal{D}_{t}$, while data from previous sessions $\mathcal{D}_0, \dots, \mathcal{D}_{t-1}$ is entirely unavailable. The objective is to maintain high performance across all previously encountered classes while effectively learning new categories from scarce data.

Formally, the dataset for each session is denoted by $\mathcal{D}_t = \{(x_i^t, y_i^t)\}_{i=1}^{M_t}$, where $M_t$ represents the total number of samples in the $t$-th session. The label spaces between different sessions are strictly disjoint: $\mathcal{Y}_s \cap \mathcal{Y}_t = \emptyset$ for any $s \neq t.$ The initial session $\mathcal{S}_{0}$ serves as the base session with a sufficient number of training samples. Subsequent sessions $\mathcal{S}_t$ $(t \geq 1)$ are typically configured as $N-$way $K-$shot problems, meaning that each new session introduces $N$ novel classes, with only $K$ examples per class. A standard FSCIL model is generally composed of a feature embedding module $f_{\theta}$ and a classification module $g_\omega$, parameterized by $\theta$ and $\omega$, respectively. For a test input $x$ from any learned session, the model predicts the label as $\hat{y} = g_\omega(f_\theta(x))$.

\subsection{Prompt-based approaches}
\label{Prompt-Based Approaches}
The core of the ViT \cite{ViT} architecture is the multi-head self-attention (MHSA) mechanism \cite{Attention_all}, which enables the model to dynamically weigh the importance of different image patches within a sequence. Formally, for an input sequence $x$, the mechanism first projects it into queries, keys, and values by using learned weight matrices $W_q$, $W_k$, and $W_v$:
\begin{equation}
Q = x W_{q},
K = x W_{k},
V = x W_{v},
\end{equation}
which are then split into $h$ heads ($Q_h, K_h, V_h$). The output of each head is computed as follows:
\begin{equation}
Z_h = \text{softmax}\left(\frac{Q_h K_h^T}{\sqrt{d_k}}\right) V_h,
\end{equation}
which allows the model to consider information from different representation subspaces. Here, $d_k$ corresponds to the dimensionality of the key vectors in each attention head. The outputs of all heads are concatenated and linearly projected by $W^O$ to form the final output $O$:
\begin{equation}
O = \text{concat}(Z_1, Z_2, \cdots, Z_h) W^O.
\end{equation}

Prompt tuning leverages the MHSA mechanism of ViT by prepending a set of learnable prompt tokens $P^l \in \mathbb{R}^{L_p \times D}$ to the input sequence of the $l$-th transformer block:
\begin{equation}
    h^{l}_{prompted} = \left[ P^{l}; h^{l} \right],
\end{equation}
where $\left[ \cdot; \cdot \right]$ denotes concatenation along the sequence dimension and $ h^{l} \in \mathbb{R}^{L_{h}\times D} $ represents the original input sequence to the $l$-th transformer block, which typically consists of the image patch embeddings and the CLS token. These prompts, which serve as task-specific context, interact with the patch and class tokens within the MHSA layers to steer the computation of the frozen pretrained backbone $f_{\theta}$ without updating its parameters $\theta$. 

\subsection{Prototype networks}
\label{Prototype networks}
Prototypical networks \cite{Prototypical} provide an efficient metric-based approach for few-shot learning. The core idea is to represent each class by its prototype, computed as the mean feature vector of its support samples:
\begin{equation}
    w_k = \frac{1}{N_k} \sum_{y_i=k} f_{\theta}(x_i),
\end{equation}
where $f_{\theta}(x_i)$ denotes the feature embedding, which  is typically derived from the CLS token representation of the ViT. These prototypes $w_k$ directly form the weights $W = [w_0, w_1, \ldots, w_k] $ of a cosine classifier. Classification is performed via a softmax function over the similarities between the input embedding and all prototypes:
\begin{equation}
    P(y = k\mid x) = \frac{\exp(\text{sim}(f_{\theta}(x), w_k))}{\sum_j \exp(\text{sim}(f_{\theta}(x), w_j))}.
\end{equation}
This method offers inherent advantages for FSCIL: new class prototypes can be directly appended to the prototype matrix to enable natural incremental learning; the old prototypes remain fixed during incremental sessions, effectively mitigating catastrophic forgetting; and the need to learn a large classification head is eliminated, resulting in high parameter efficiency.

\section{The proposed method}
\label{Method}
\begin{figure}[t]
\centering
\includegraphics[width=1.0\textwidth]{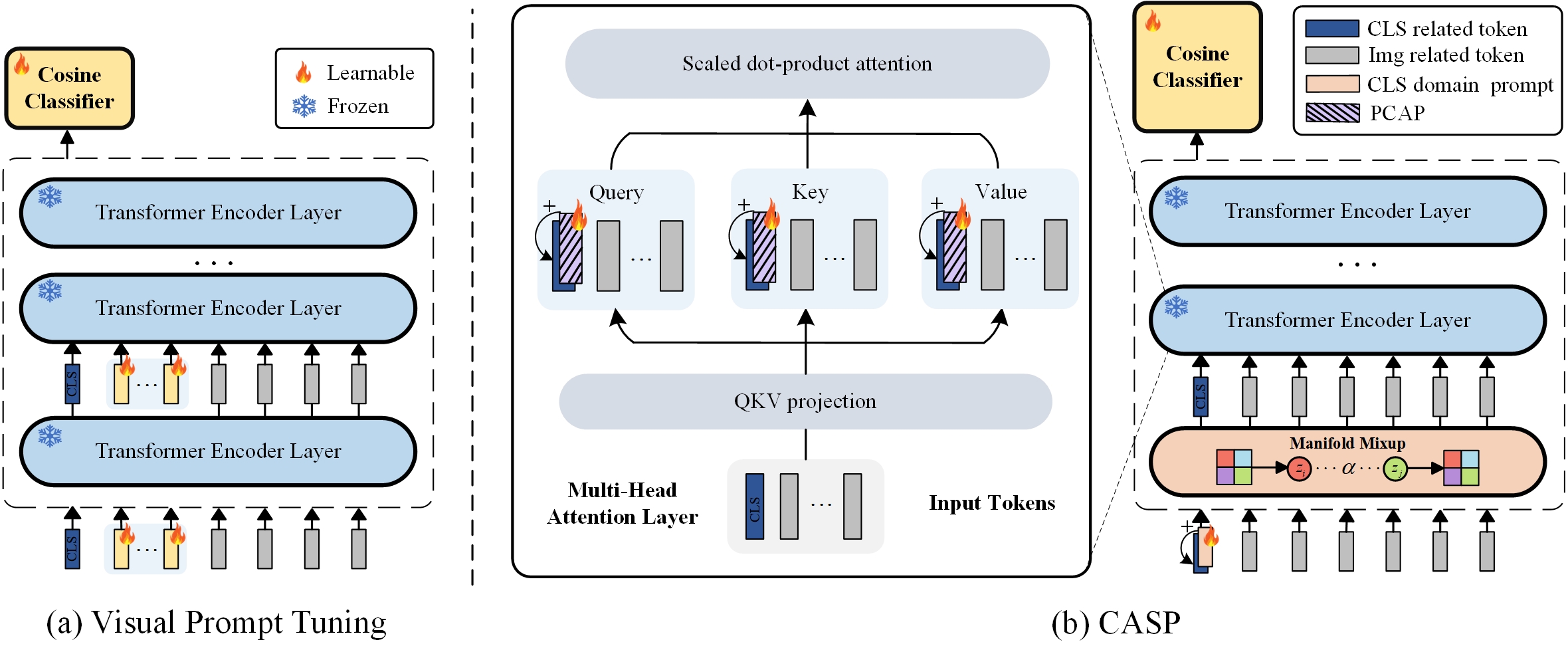}
\caption{CASP Architecture. The CASP framework introduces class-shared bias parameters into CLS token projections and employs feature space Mixup, enabling dynamic self-attention adjustment and generalization without full fine-tuning.}
\label{model}
\end{figure}

The special role of the CLS token in the ViT architecture \cite{ViT} indicates that enhancing the discriminability and generalization capability of the CLS token representation is crucial for effective FSCIL. To this end, we propose a novel additive prompt tuning framework that strengthens the representational capacity of the CLS token from three complementary aspects: attention-guided prompts for the CLS token, cross-domain additive prompts for the CLS token, and feature space preservation for future classes. This comprehensive approach enhances the ability of the ViT model to transfer pretrained knowledge and generalize it to unknown categories. The overall architecture of our method is illustrated in Fig. \ref{model}.

\subsection{CLS token attention-guided prompts}
\label{CLS Attention-Guided Prompts}
The self-attention mechanism is fundamental to ViT, as it governs the contextualization process whereby the CLS token aggregates information from all of the patch tokens. To proactively guide and refine this critical process, we introduce a set of CLS Token Attention-Guided Prompts (CAGP) that directly modulate the key constituents of the attention mechanism for the CLS token.

Formally, within each multi-head self-attention (MSA) layer, we inject learnable prompt vectors into the query, key, and value projections of the CLS token. Let the input token sequence be denoted by $X = [x_{cls}; X_{patch}] \in \mathbb{R}^{(N+1) \times D}$, where $x_{cls}$ is the CLS token. The standard query, key, and value projections for the CLS token are obtained through linear transformations:
\begin{equation}
    q_{cls} = x_{cls} W_q, \quad k_{cls} = x_{cls} W_k, \quad v_{cls} = x_{cls} W_v.
\end{equation}
To adapt the attention dynamics specifically for the downstream task, we augment the CLS token projections with dedicated prompt vectors:
\begin{equation}
    \tilde{q}_{cls} = q_{cls} + p_q, \\
    \tilde{k}_{cls} = k_{cls} + p_k, \\
    \tilde{v}_{cls} = v_{cls} + p_v,
\end{equation}
where $p_q, p_k, p_v \in \mathbb{R}^D$ are the learnable additive prompts. This operation effectively shifts the position of the CLS token in the query, key, and value spaces, thereby steering its attention relationships with the patch tokens towards a better configuration.

\subsection{Perturbed attention of the CLS token}
\label{Perturbed Attention of CLS Token}
Merely introducing learnable parameters into the attention mechanism may lead to overfitting on the base classes. To enhance the generalization capability of the prompt vectors to future categories, we apply stochastic perturbation to the CLS token attention during training, preventing the prompt module from overfitting to noise in the base session. This method is referred to as Perturbed CLS Token Attention-Guided Prompts (PCAP). Specifically, before the prompt vectors are incorporated into the attention computation during training, dropout \cite{SRIVASTAVA2014} is applied to improve the model's generalization performance:
\begin{equation}
    \begin{aligned}
    p'_q = Dropout(p_q), \\
    p'_k = Dropout(p_k), \\
    p'_v = Dropout(p_v),
    \end{aligned}
\end{equation}
where the dropout operation independently masks the elements of each prompt vector with a predefined probability. This regularization technique not only discourages overreliance on specific prompt features but also encourages the learning of a more resilient and transferable representation for the CLS token, thereby facilitating effective knowledge generalization to novel categories.

\subsection{CLS domain adaptation prompt}
\label{CLS Domain Adaptation Prompts}
Recent studies \cite{Zou2024} have revealed that the standard CLS token in the Vision Transformer tends to absorb and retain domain-specific information from the pretraining dataset (e.g., ImageNet-1K), which often manifests as low-frequency components in the Fourier spectrum. While beneficial for the source domain, this ingrained domain bias can hinder the model's adaptability and generalizability to novel downstream tasks.

To address this limitation, we propose the CLS Domain Adaptation Prompt (CDAP), which learns an additive and task-specific vector that actively counteracts and neutralizes the source-domain information embedded within the pretrained CLS token. To mitigate the domain-specific bias inherent in the pretrained CLS token, we introduce a learnable vector $p_d \in \mathbb{R}^D$, which is designed explicitly for domain information decoupling. This prompt is introduced only at the input stage of the transformer encoder. Specifically, $p_d$ is added to the original CLS token before the first encoder layer, producing an adapted token that serves as the initial representation for subsequent processing. The modified CLS token is computed as follows:
\begin{equation}
    \tilde{x}_{cls}^{(0)} = x_{cls}^{(0)} + p_d,
\end{equation}
where $x_{cls}^{(0)}$ denotes the original pretrained CLS token and $\tilde{x}_{cls}^{(0)}$ represents the domain-adapted token input to the first transformer layer. During base-session training, $p_d$ is optimized to counteract the source-domain characteristics embedded in $x_{cls}^{(0)}$. This additive adjustment effectively decouples spurious domain-specific correlations, resulting in a more flexible and neutral initial representation. Consequently, the attention mechanism can more effectively integrate features from novel target-domain patches, significantly enhancing the model's cross-domain generalization capability.

\subsection{Manifold token mixup for feature augmentation}
To enhance the model's generalization capability and explicitly reserve the representation space for unseen future categories, we introduce a feature-level augmentation strategy during the base training phase. Specifically, we perform mixing in the token space of the ViT. This space retains rich structural and semantic information from the input patches while being sufficiently abstract to facilitate meaningful interpolation, thereby synthesizing robust features that mimic potential novel categories. 

Our proposed Manifold Token Mixup (MTM) is integrated into the base training phase. For an input batch of images $X$, we extract shallow tokenized representations by using an initial feature extractor $f_{pre}(\cdot)$, which comprises the early layers of the ViT encoder:
\begin{equation}
    Z = f_{pre}(X),
\end{equation}
where $Z$ denotes the feature representation that includes both the adapted CLS token and all of the patch tokens. We stochastically mix these feature representations within the mini-batch. A random permutation index $idx$ is generated, and a mixing coefficient $\beta$ is sampled from a beta distribution. The mixed feature manifold $\tilde{Z}$ is obtained via linear interpolation: 
\begin{equation}
    \tilde{Z} = \beta \cdot Z + (1 - \beta) \cdot Z[idx].
\end{equation}
The mixed feature $\tilde{Z}$ is subsequently propagated through a deeper feature extractor $f_{post}(\cdot)$, which consists of the later layers of the ViT encoder, to obtain the output logits $\tilde{Y}$:
\begin{equation}
    \tilde{Y} = f_{post}(\tilde{Z}).
\end{equation}
Similarly, a soft label $\tilde{T}$ is constructed by using the same $\beta$ to blend the original and permuted one-hot labels: 
\begin{equation}
    \tilde{T} = \beta \cdot T + (1 - \beta) \cdot T[idx].
\end{equation}

The model's optimization objective is to minimize a composite loss function. This is achieved by combining the standard cross-entropy loss $\mathcal{L}_{ce}$ on the original data with a soft cross-entropy loss $\mathcal{L}_{mix}$ applied to the mixed feature representations. The latter is defined as:
\begin{equation}
    \mathcal{L}_{\text{mix}} = -\tilde{T} \cdot \log(softmax(\tilde{Y})),
\end{equation}
which measures the discrepancy between the predictions $\tilde{Y}$ for the mixed features and their corresponding soft labels $\tilde{T}$. The total loss function of the proposed method is: 
\begin{equation}
    \mathcal{L}_{total} = \mathcal{L}_{ce} + \lambda_{mix} \mathcal{L}_{mix},
\end{equation}
with $\lambda_{mix}$ controlling the contribution of the mixing loss.

\begin{table*}[t]
    \caption{10-way 5-shot performance on the CUB200 dataset with the pretrained ViT-B/16-1K backbone.}
    \label{tab:cub200_results}
    \centering
    \resizebox{1.0\textwidth}{!}{%
    \begin{tabular}{lcccccccccccr}
    \hline
    \multicolumn{1}{c}{\multirow{2}{*}{Method}} & \multicolumn{11}{c}{Acc. in each session (\%)} & \multicolumn{1}{c}{\multirow{2}{*}{Avg (\%)}} \\ \cline{2-12}
    \multicolumn{1}{c}{} & 0 & 1 & 2 & 3 & 4 & 5 & 6 & 7 & 8 & 9 & 10 & \multicolumn{1}{c}{} \\ \hline
    L2P \cite{L2P} & 82.4 & 81.2 & 79.0 & 76.8 & 76.2 & 74.7 & 74.1 & 74.1 & 72.7 & 73.0 & 73.6 & 76.2 \\ 
    DualP \cite{DualPrompt} & 83.5 & 82.2 & 80.9 & 79.5 & 78.6 & 77.0 & 76.3 & 77.0 & 75.7 & 76.1 & 76.5 & 78.5 \\ 
    Coda-P \cite{CODA-Prompt} & 79.6 & 78.1 & 76.4 & 75.6 & 75.0 & 73.1 & 72.5 & 72.8 & 72.0 & 72.4 & 72.9 & 74.6 \\ \hline
    TEEN \cite{TEEN} & \underline{88.8} & 86.2 & 85.5 & 83.0 & 83.0 & 81.7 & 81.5 & 79.7 & 79.9 & 79.5 & 80.0 & 82.6 \\ 
    PriViLege \cite{PKT} & 82.3 & 81.3 & 80.5 & 77.8 & 77.8 & 76.0 & 75.7 & 76.0 & 75.2 & 75.2 & 75.1 & 77.5 \\
    LGSP \cite{LGSP} & 85.7 & 84.3 & 83.2 & 81.3 & 81.8 & 80.3 & 79.9 & 80.1 & 79.2 & 79.7 & 79.7 & 81.4 \\
    ASP \cite{ASP} & 87.1 & 86.0 & 84.9 & 83.4 & 83.6 & 82.4 & 82.6 & 83.0 & 82.6 & 83.0 & 83.5 & 83.8 \\
    SEC \cite{SEC} & 87.5 & \underline{86.6} & \underline{85.5} & \underline{84.6} & \underline{84.9} & \underline{84.0} & \underline{83.7} & \underline{84.2} & \underline{83.8} & \underline{84.0} & \underline{84.8} & \underline{84.9} \\ \hline
    $\mathbf{CASP\ (Ours)}$ & $\mathbf{89.4}$ & $\mathbf{87.9}$ & $\mathbf{87.4}$ & $\mathbf{86.8}$ & $\mathbf{86.8}$ & $\mathbf{85.0}$ & $\mathbf{85.0}$ & $\mathbf{85.4}$ & $\mathbf{85.3}$ & $\mathbf{85.6}$ & $\mathbf{85.8}$ & $\mathbf{86.4}$ \\ \hline
    \end{tabular}%
    }
    
\end{table*}

\begin{table*}[t]
    \caption{5-way 5-shot performance on the CIFAR100 dataset with
    the pretrained ViT-B/16-1K backbone.}
    \label{tab:results_CIFAR100}
    \centering
    
    \resizebox{0.9\textwidth}{!}{
    \begin{tabular}{@{}lcccccccccc@{}}
    \toprule
    \multirow{2}{*}{Method} & \multicolumn{9}{c}{Acc. in each session (\%)} & \multirow{2}{*}{Avg (\%) } \\ \cmidrule(r){2-10}
     & 0 & 1 & 2 & 3 & 4 & 5 & 6 & 7 & 8 &  \\ \midrule
    L2P \cite{L2P} & 84.7 & 82.3 & 80.1 & 77.5 & 77.0 & 76.0 & 75.6 & 74.1 & 72.3 & 77.7 \\
    DualP \cite{DualPrompt} & 86.0 & 83.6 & 82.9 & 80.2 & 80.6 & 80.2 & 80.5 & 79.0 & 77.4 & 81.1 \\
    Coda-P \cite{CODA-Prompt} & 86.0 & 83.6 & 81.6 & 79.2 & 79.1 & 78.5 & 78.3 & 77.0 & 75.4 & 79.8 \\ \midrule
    TEEN \cite{TEEN} & \underline{92.9} & 90.2 & 88.4 & 86.8 & 86.4 & 86.0 & 85.8 & 85.1 & 84.0 & 87.3 \\
    PriViLege \cite{PKT} & 90.9 & 89.4 & 89.0 & 87.6 & 87.9 & 87.4 & 87.6 & 87.2 & 86.1 & 88.1 \\
    ASP \cite{ASP} & 92.2 & 90.7 & 90.0 & 88.7 & 88.7 & 88.2 & 88.2 & 87.8 & 86.7 & 89.0 \\ 
    SEC \cite{SEC} & 92.0 & \underline{90.8} & \underline{90.6} & \underline{89.1} & \underline{89.2} & \underline{89.1} & \underline{89.2} & \underline{88.7} & \underline{87.5} & \underline{89.5} \\ \midrule
    $\mathbf{CASP\ (Ours)}$ &  $\mathbf{93.4}$ & $\mathbf{91.8}$ & $\mathbf{91.0}$ & $\mathbf{89.7}$ & $\mathbf{89.8}$ & $\mathbf{89.5}$ & $\mathbf{89.5}$ & $\mathbf{89.3}$ & $\mathbf{88.1}$ & $\mathbf{90.2}$ \\ \bottomrule
    \end{tabular}%
    }
    
    \end{table*}

\begin{table*}[t]
    \caption{10-way 5-shot performance on the ImageNet-R dataset with the pretrained ViT-B/16-1K backbone.}
    \label{tab:results_ImageNet-R}
    \centering
    \resizebox{1.0\textwidth}{!}{%
    \begin{tabular}{lcccccccccccr}
    \hline
    \multicolumn{1}{c}{\multirow{2}{*}{Method}} & \multicolumn{11}{c}{Acc. in each session (\%)} & \multicolumn{1}{c}{\multirow{2}{*}{Avg (\%)}} \\ \cline{2-12}
    \multicolumn{1}{c}{} & 0 & 1 & 2 & 3 & 4 & 5 & 6 & 7 & 8 & 9 & 10 & \multicolumn{1}{c}{} \\ \hline
    L2P \cite{L2P} & 73.9 & 70.9 & 69.3 & 65.9 & 64.0 & 62.6 & 60.1 & 59.5 & 59.0 & 58.2 & 56.8 & 63.7 \\ 
    DualP \cite{DualPrompt} & 71.5 & 69.0 & 69.0 & 67.4 & 66.6 & 65.9 & 64.1 & 64.0 & 64.0 & 63.4 & 62.5 & 66.1 \\ 
    Coda-P \cite{CODA-Prompt} & 82.8 & 80.1 & 78.6 & 75.3 & 73.1 & 71.8 & 70.0 & 69.2 & 68.8 & 67.4 & 66.4 & 73.0 \\ \hline
    TEEN \cite{TEEN} & $\mathbf{84.6}$ & 76.7 & 68.8 & 67.6 & 64.3 & 60.6 & 58.3 & 56.1 & 56.1 & 54.7 & 54.9 & 63.9 \\ 
    PriViLege \cite{PKT} & 80.0 & 77.3 & 77.0 & 73.8 & 72.8 & 72.3 & 71.3 & 69.7 & 69.5 & 67.7 & 66.5 & 72.5 \\
    ASP \cite{ASP} & 83.3 & 80.4 & 79.6 & 77.0 & 75.6 & 74.7 & 73.0 & 72.1 & 71.9 & 70.9 & 69.7 & 75.3 \\
    SEC-F \cite{SEC} & 82.9 & \underline{80.5} & \underline{79.8} & \underline{77.5} & \underline{76.7} & \underline{76.3} & \underline{74.8} & \underline{74.6} & \underline{74.1} & \underline{73.8} & \underline{73.0} & \underline{76.7} \\ \hline
    $\mathbf{CASP\ (Ours)}$ & \underline{83.7} & $\mathbf{80.8}$ & $\mathbf{80.5}$ & $\mathbf{78.2}$ & $\mathbf{77.4}$ & $\mathbf{76.5}$ & $\mathbf{75.4}$ & $\mathbf{75.2}$ & $\mathbf{75.1}$ & $\mathbf{74.5}$ & $\mathbf{73.8}$ & $\mathbf{77.4}$ \\ \hline
    \end{tabular}%
    }
    
\end{table*}

\section{Experiments}
\label{Experiments}

\subsection{Experimental configuration}
\label{Experimental configuration}
In this section, we provide a detailed description of the validation process for the proposed method. The experimental data are sourced from three widely used computer vision datasets: CUB200 \cite{CUB2011}, CIFAR100 \cite{CIFAR100}, and ImageNet-R \cite{Hendrycks2021}. Throughout the study, we adhere strictly to the standard experimental configurations \cite{ASP,SEC} and splitting protocols in the field, dividing the datasets into base and incremental classes to accommodate the FSCIL scenario. The base session evaluation focuses on the model's performance on the initial classes, while the incremental sessions assess the model's adaptability when new classes are introduced. To comprehensively reflect the model's performance, we pay particular attention to four key metrics: the base session accuracy ($A_B$), the last session accuracy ($A_L$), the last session's new class accuracy ($A_N$), and the average accuracy across all sessions ($A_{avg}$). We conduct 5 simulations under different random seeds and report the averages.

\subsection{Baselines and training details}
For comparison, several recently proposed FSCIL algorithms, including PriViLege \cite{PKT}, ASP \cite{ASP}, and SEC \cite{SEC}, are selected as baseline methods. All methods utilize a ViT-B/16-1K \cite{ViT} model pretrained on ImageNet-1K as the backbone network. In our approach, the CAGP is embedded into all layers of the backbone, whereas for CDAP, only the initial ViT CLS token is fine-tuned. For the CUB200 dataset, the MTM module is applied to the first layer with $\lambda_{mix}$ = 0.5. For the other two datasets, the MTM is applied to the sixth layer with $\lambda_{mix}$ = 0.05. These hyperparameters are determined via a simple grid search.

During training, we adopt the Adam optimizer with a cosine annealing scheduler, using an initial learning rate of $1 \times 10^{-2}$ and a batch size of 64. All experiments are conducted on an RTX 4090 GPU. For the base session, training is carried out for 30 epochs. We observe that incremental fine-tuning exhibits unstable performance across different datasets, showing inconsistent gains that involve slight improvements on the CUB200 dataset but performance degradation on the others. To ensure the robustness of the experimental results and fairness of the evaluation across different tasks, we ultimately employ a unified strategy of fine-tuning only in the base session for all of the experiments.

\subsection{Main experimental results}
\label{Main experimental results}
\begin{figure}[t]
    \centering
    \includegraphics[width=1.0\textwidth]{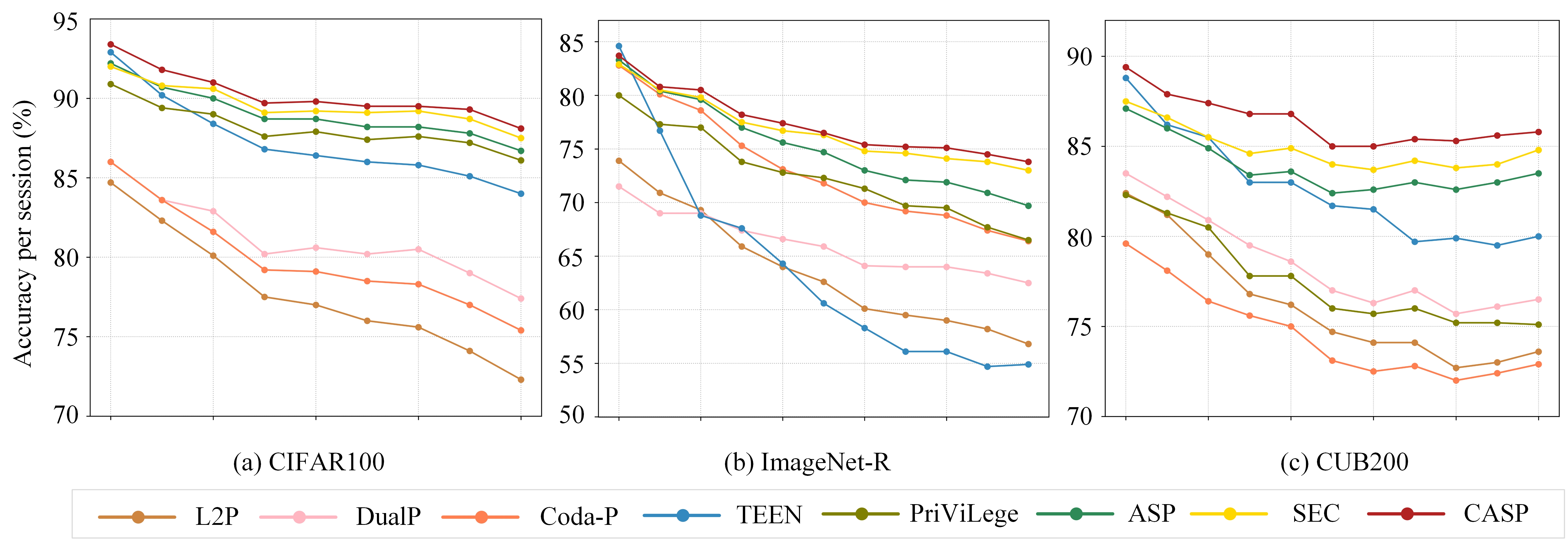}
    \caption{SOTA comparisons on the CIFAR100, ImageNet-R, and CUB200 benchmarks.}\label{fig1}
    \label{sota_performance} 
\end{figure}

We evaluate the proposed CASP method by comparing it with existing approaches on three benchmark datasets: CUB200 (Table \ref{tab:cub200_results}), CIFAR100 (Table \ref{tab:results_CIFAR100}), and ImageNet-R (Table \ref{tab:results_ImageNet-R}). The experimental results demonstrate that CASP achieves state-of-the-art performance across all three datasets, with particularly notable gains on the fine-grained CUB200 dataset. As summarized in Table \ref{tab:cub200_results}, CASP outperforms the second-best method by 1.5\% in terms of average accuracy ($A_{avg}$) and 1\% in terms of the final session accuracy ($A_L$).  These comparative results are also intuitively displayed in Fig. \ref{sota_performance}, which clearly illustrates the performance advantage of our method across incremental sessions. On CIFAR100, CASP also exhibits strong performance, exceeding the second-best approach by 0.7\% in terms of $A_{avg}$
 and 0.6\% in terms of $A_L$. Notably, the proposed method requires no fine-tuning during the incremental phases, resulting in a more streamlined and practical architecture.

\subsection{Parameter efficiency comparison}
\label{Parameter efficiency comparison}
To further evaluate the practical efficiency of our method, we compare the parameter efficiency of CASP against that of previous FSCIL approaches on the CUB200 dataset using the ViT-B/16-1K backbone. As summarized in Table \ref{tab:casp_comparison}, our method achieves a remarkable reduction in the number of trainable parameters. Specifically, CASP introduces only 0.027 MB of tunable weights within the backbone itself. The total trainable parameters, including the classifier, amount to only 0.1 MB. This allows CASP to require significantly fewer training resources than the closest competitor, SEC-F (1.7 MB) \cite{SEC}, while maintaining a minimal total parameter footprint comparable to that of a fully frozen backbone. These results underscore the exceptional parameter efficiency of the proposed approach.

\begin{table}[t]
  \centering
  \caption{Resource comparison on the CUB200 dataset with the ViT-B/16-1K backbone.}
  \label{tab:casp_comparison}
  \begin{tabular}{|l|c|c|}
    \hline
    \textbf{Methods} & \textbf{trainable params (MB)} & \textbf{Total params (MB)} \\
    \hline
    PriViLege \cite{PKT} & 16.3 & 102.3 \\ \hline
    ASP \cite{ASP} & 3.0 & 103.0 \\ \hline
    SEC \cite{SEC} & 2.8 & 102.8 \\ \hline
    SEC-F \cite{SEC} & 1.7 & 101.7 \\ \hline
    CASP (ours) & \textbf{0.1} & \textbf{100.1}  \\
    \hline
  \end{tabular}
\end{table}

\definecolor{darkred}{RGB}{178, 34, 34} 
\begin{table}[t]
    \caption{Ablation studies on the CUB200 dataset.}
    \label{tab:ablation_cUB200}
    \centering
    
    \resizebox{0.8\textwidth}{!}{
    \begin{tabular}{@{}cccccccc@{}}
    \toprule
    $CAGP$ & $PCAP$ & $CDAP$ & $MTM$  & $A_B$ (\%) & $A_N$ (\%) & $A_L$ (\%) & $A_{avg}$ (\%) \\ \midrule
    \ding{55}       & \ding{55}        & \ding{55}     &\ding{55}          & 85.2      & 16.4      & 28.2      & 30.6   \\
    \textcolor{darkred}{\checkmark}        & \ding{55}  &\ding{55}      & \ding{55}        & 88.5      & 84.1      & 84.8      & 85.4 \\
    \textcolor{darkred}{\checkmark}        & \textcolor{darkred}{\checkmark}  &\ding{55}       & \ding{55}       & 88.9      & 84.8      & 85.2      & 85.9   \\

    \textcolor{darkred}{\checkmark}        & \textcolor{darkred}{\checkmark}        & \textcolor{darkred}{\checkmark} &\ding{55}       & \underline{89.2}      & \underline{85.0}      & \underline{85.6} & \underline{86.1}   \\
    \textcolor{darkred}{\checkmark}       & \textcolor{darkred}{\checkmark}          & \textcolor{darkred}{\checkmark}  &\textcolor{darkred}{\checkmark}       & \textbf{89.4}      & \textbf{85.3}      & \textbf{85.8}      & \textbf{86.4}   \\ \bottomrule
    \end{tabular}
    }
\end{table}


\subsection{Ablation studies}
\label{Ablation Studies}
We conduct ablation studies on the CUB200 to validate the effectiveness of our proposed method. Table \ref{tab:ablation_cUB200} presents a performance comparison with incremental incorporation of each component after employing cosine prototype classifiers and a fully fine-tuned pretrained ViT as baselines. The results demonstrate that the progressive introduction of each component leads to consistent performance improvements. When only the CAGP is introduced, the model shows a remarkable increase in incremental learning, with the new-class accuracy ($A_N$) increasing substantially from 16.4\% to 84.1\%, and further integration of the PCAP, CDAP, and MTM modules leads to steady growth across all of the metrics. The best performance is achieved when all of the components are combined, yielding a final session accuracy ($A_L$) of 85.8\% and an average accuracy ($A_{avg}$) of 86.4\%. These results indicate that each component contributes at different levels and exhibits strong complementarity.

\begin{figure}[t]
    \centering
    \includegraphics[width=1.0\textwidth]{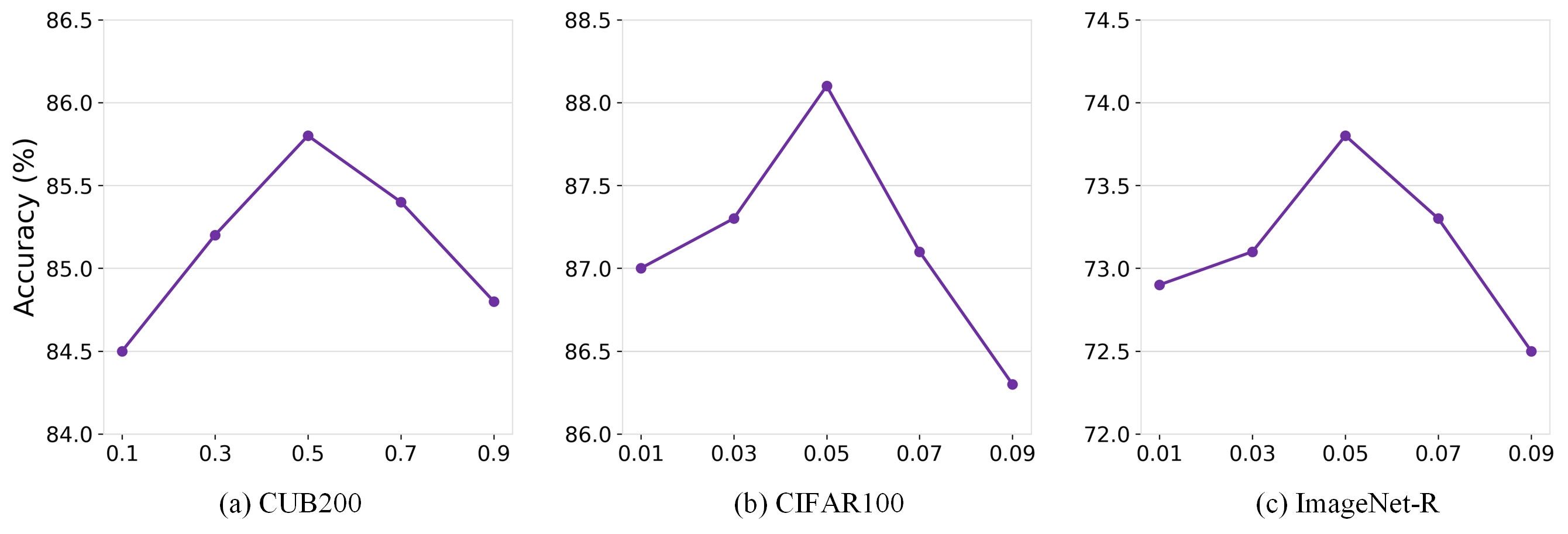}
    \caption{Sensitivity study of the Manifold Token Mixup hyperparameter $\lambda_{mix}$,
    illustrating the model's $A_{L}$ across varying values of $\lambda_{mix}$. }
    \label{mixhyperparameter} 
\end{figure}

\begin{figure}[t]
    \centering
    \includegraphics[width=1.0\textwidth]{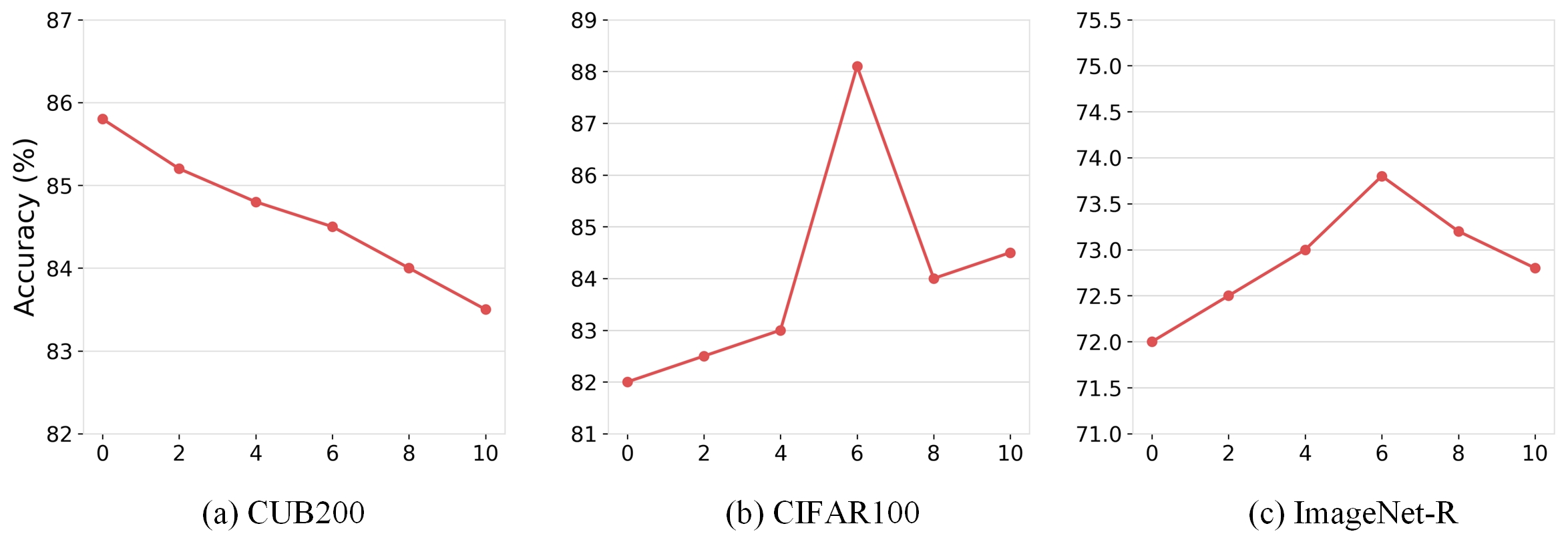}
    \caption{ Performance trend of $A_L$ plotted against the depth of the layer where the MTM is applied.}
    \label{layerhyperparameter}
\end{figure}

\subsection{Analysis of the hyperparameters}
\label{Analysis of hyperparameter}
To determine the optimal model configuration, we conduct a systematic hyperparameter sensitivity analysis. As illustrated in Fig. \ref{mixhyperparameter}, the impact of the MTM hyperparameter $\lambda_{mix}$ on the final session accuracy ($A_L$) is evaluated. The results indicate that for all three datasets, there exists a distinct optimal value of $\lambda_{mix}$ 
 that maximizes model performance. The effects of applying the MTM module at different depths of the network are further examined in Fig. \ref{layerhyperparameter}. The performance trend reveals that the optimal layer for the MTM is closely related to dataset characteristics: for the fine-grained CUB200 dataset, applying the MTM at shallower layers yields better results, whereas for the more complex CIFAR100 and ImageNet-R datasets, integrating the MTM into deeper layers results in greater performance gains. These findings emphasize the importance of adapting the model structure according to the specific task.

\begin{figure}[t]
    \centering
    \includegraphics[width=1.0\textwidth]{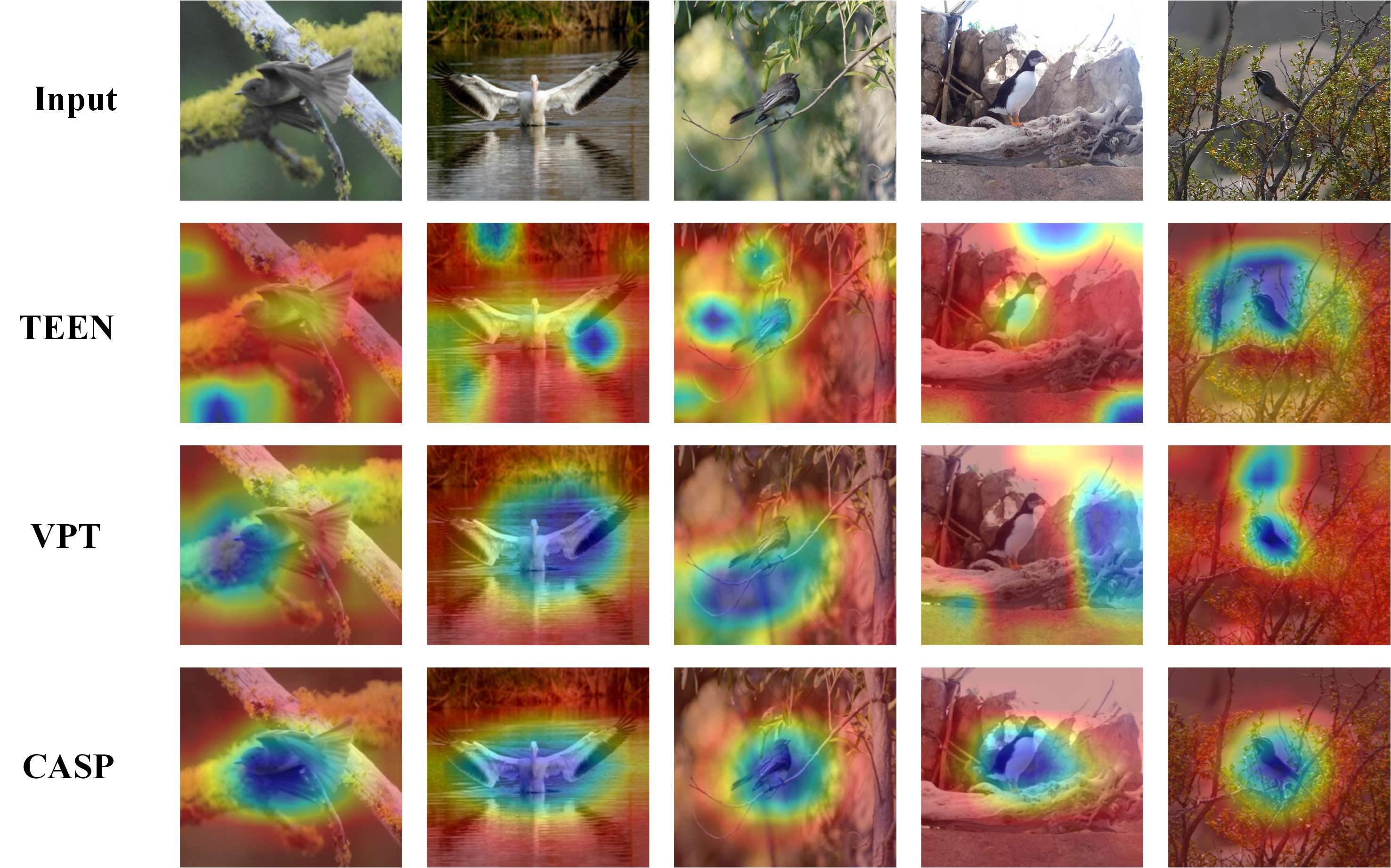}
    \caption{Visualization comparison of the class activation maps on images of birds. TEEN: Training-free prototype calibration; VPT: Visual prompt tuning; CASP: Our proposed method.}
    \label{cam}
\end{figure}

\begin{figure}[t]
    \centering
    \includegraphics[width=1.0\textwidth]{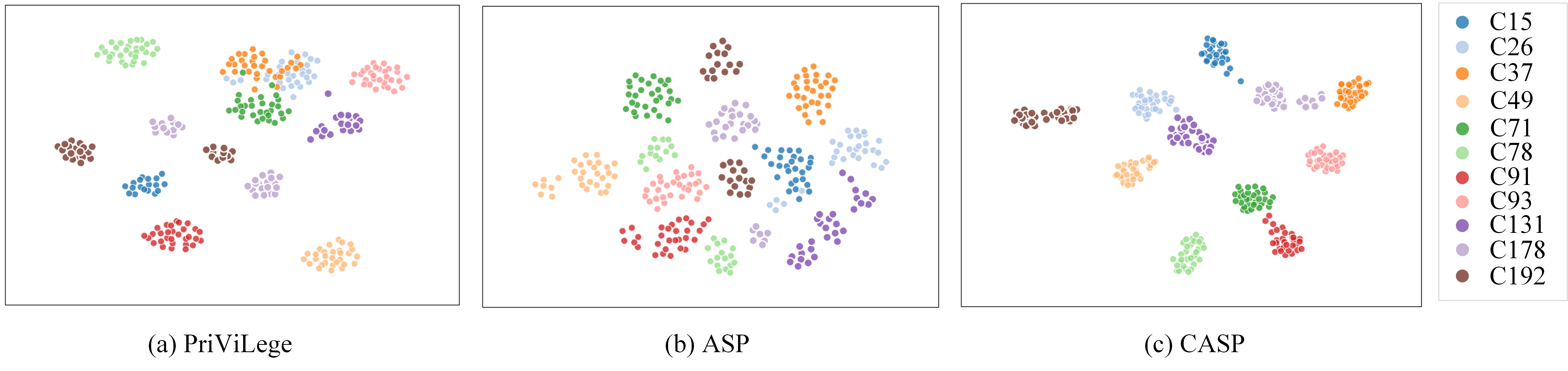}
    \caption{T-SNE visualization of the learned representations on the CUB200 dataset. }
    \label{t-sne}
\end{figure}

\subsection{Visualization}
\label{Visualization}

To qualitatively evaluate the ability of different methods to capture discriminative features, we present a visualization comparison of 
the class activation maps (CAMs) \cite{DeepFeaturesDiscriminativeLocalization} on bird images from the CUB200 dataset in Fig. \ref{cam}. The visualization clearly contrasts the activation patterns of our CASP method with those of two established baselines, TEEN \cite{TEEN} and the VPT \cite{VPT}. As the results show, the attention maps generated by TEEN and the VPT are often diffuse and tend to be activated over broad background regions or less semantically relevant parts of the bird. In contrast, the heatmaps produced by our CASP method are significantly more focused and precise, consistently highlighting the most discriminative regions, such as the head and torso of the birds, while effectively suppressing responses from the cluttered background. This side-by-side comparison provides intuitive evidence that CASP learns more semantically meaningful representations, which aligns with its superior quantitative performance shown in the previous sections.

Furthermore, to validate the effectiveness of the learned feature representations in the embedding space, we provide t-SNE visualizations \cite{TSNEVisualization} of the feature distributions on the CUB200 dataset in Fig. \ref{t-sne}. The visualization compares the feature separability of (a) PriViLege, (b) ASP, and (c) our proposed CASP method. As clearly demonstrated, the feature clusters generated by CASP exhibit more compact intraclass distributions and larger interclass margins than those of the other methods. While PriViLege and ASP show varying degrees of class overlap and dispersed clusters, CASP produces well-separated groupings with minimal ambiguity. This enhanced separability in the latent space provides additional evidence that CASP learns more discriminative representations, effectively mapping different classes to distinct regions of the feature space. Together with the CAM visualization results, these findings consistently demonstrate that our method excels at both capturing semantically meaningful local features and generating globally separable representations.

\section{Conclusion}
\label{Conclusion}
To enhance feature generalization and discriminability in FSCIL, we propose a prompt-based framework named CASP. This framework employs a symmetric prompt injection mechanism that directs the model's attention to critical features while incorporating attention perturbation and Manifold Token Mixup strategies to further improve robustness. Experimental results on multiple benchmark datasets, including CUB200, CIFAR100, and ImageNet-R, demonstrate that CASP achieves optimal performance during the incremental learning phase without the need for fine-tuning while significantly reducing the number of trainable parameters and exhibiting exceptional generalizability and stability. In the future, we will explore FSCIL in multimodal scenarios and extend the core mechanisms of CASP to other learning tasks that require efficient representation transfer.

\end{document}